\documentclass{bmvc2k}

\usepackage{multirow}
\usepackage{bm}
\usepackage{amsmath, amsthm, amsfonts}
\usepackage{booktabs}
\usepackage{bbding}  
\usepackage[inline]{enumitem}
\usepackage{wrapfig}  
\usepackage{tikz}
\usepackage{pgfplots}

\newcommand{\bp}{\ensuremath{\bm{p}}}

\newcommand{\bh}{\ensuremath{\bm{h}}}

\newcommand{\bb}{\ensuremath{\bm{b}}}
\newcommand{\bs}{\ensuremath{\bm{s}}}

\newcommand{\bc}{\ensuremath{\bm{c}}}
\newcommand{\bq}{\ensuremath{\bm{q}}}
\newcommand{\bl}{\ensuremath{\bm{l}}}

\newcommand{\bS}{\ensuremath{\bm{S}}}

\newcommand{\bP}{\ensuremath{\bm{P}}}

\newcommand{\bI}{\ensuremath{\bm{I}}}

\definecolor{colorblue}{rgb}{0.29, 0.59, 0.82}
\definecolor{colorred}{rgb}{0.9, 0.17, 0.31}
\definecolor{colorred}{rgb}{0.9, 0.17, 0.31}
\definecolor{coolgrey}{rgb}{0.55, 0.57, 0.67}
\definecolor{colorwhite}{rgb}{1, 1, 1}

\let\oldTheta\theta
\renewcommand{\theta}[1]{\ensuremath{\oldTheta_{#1}}}

\providecommand*{\Rset}{\mathbb{R}}  

\newcommand*{\calL}{\ensuremath{\mathcal{L}}}

\newcommand*{\calP}{\ensuremath{\mathcal{P}}}

\DeclareMathOperator*{\argmax}{argmax} 

\newcommand*{\qj}{\bm{q}_j}
\newcommand*{\eqj}{\bm{e}^{\qj}}
\newcommand*{\Eqj}{E^{\qj}}
\newcommand*{\eqjt}{\overline{\bm{e}}^{\qj}}
\newcommand*{\Eqjt}{\overline{E}^{\qj}}
\newcommand*{\ePI}{\bm{e}^{\calP_{\bm{I}}}}
\newcommand*{\EPI}{E^{\calP_{\bm{I}}}}
\newcommand*{\ePIt}{\overline{\bm{e}}^{\calP_{\bm{I}}}}
\newcommand*{\EPIt}{\overline{E}^{\calP_{\bm{I}}}}
\newcommand*{\fsim}{f_{\text{sim}}}

\newcommand*{\sent}{\mathrm{S}}
\newcommand*{\erqj}{\bm{\xi}}  

\title{Weakly-Supervised Visual-Textual Grounding with Semantic Prior Refinement}

\addauthor{Davide Rigoni}{davide.rigoni.2@phd.unipd.it}{1,2}
\addauthor{Luca Parolari}{luca.parolari@unipd.it}{1}
\addauthor{Luciano Serafini}{serafini@fbk.eu}{2}
\addauthor{Alessandro Sperduti}{alessandro.sperduti@unipd.it}{1,3}
\addauthor{Lamberto Ballan}{lamberto.ballan@unipd.it}{1}

\addinstitution{
University of Padova\\
Padova, Italy
}
\addinstitution{
Fondazione Bruno Kessler (FBK)\\
Povo, Italy
}
\addinstitution{
University of Trento \\
Trento, Italy
}

\runninghead{Rigoni et al}{Semantic Prior Refinement Model}

\pgfplotsset{compat=1.18} 
\begin{document}

\maketitle

\begin{abstract}
Using only image-sentence pairs, weakly-supervised visual-textual grounding aims to learn region-phrase correspondences of the respective entity mentions. Compared to the supervised approach, learning is more difficult since bounding boxes and textual phrases correspondences are unavailable. 
In light of this, we propose the Semantic Prior Refinement Model (SPRM), whose predictions are obtained by combining the output of two main modules. The first untrained module aims to return a rough alignment between textual phrases and bounding boxes. The second trained module is composed of two sub-components that refine the rough alignment to improve the accuracy of the final phrase-bounding box alignments.
The model is trained to maximize the multimodal similarity between an image and a sentence, while minimizing the multimodal similarity of the same sentence and a new unrelated image, carefully selected to help the most during training. Our approach shows state-of-the-art results on two popular datasets, Flickr30k Entities and ReferIt, shining especially on ReferIt with a 9.6\% absolute improvement. Moreover, thanks to the untrained component, it reaches competitive performances just using a small fraction of training examples.
\end{abstract}

\section{Introduction}
\label{sec:introduction}

Visual-textual Grounding (VG), i.e. the task of locating objects referred by natural language sentences, requires a joint understanding of both visual and textual modalities. 
Depending on the amount of annotations used during training, VG can be tackled in a different manner. 
In this work, we focus on the weakly-supervised setting~\cite{arbelle2021detector,wang2020maf,gupta2020contrastive,wang2019phrase,zhao2018weakly} in which the only available annotation refers to image-sentence pairs. 
In other words, it is only known which sentence describes each image in the dataset, but not the objects in the image referred by the textual phrases composing the sentence. 
In contrast, in a fully-supervised setting the model is trained using all the region-phrase pairs~\cite{rohrbach2016grounding,chen2018knowledge,zhang2018grounding,gupta2020contrastive,DBLP:conf/sac/RigoniSS22,DBLP:conf/iccv/KamathSLSMC21}, which in practice is a difficult and very expensive annotation to collect.

To this end, we propose a simple model referred as to Semantic Prior Refinement Model (SPRM), whose predictions are obtained by combining two modules: 
\begin{enumerate*}[label=(\roman*)]
    \item the first, which \textbf{does not require training}, for each textual phrase returns a rough alignment with a candidate object bounding box, 
    \item while the second, composed by two \textbf{trained} sub-components, refines the rough alignments in the final phrase-bounding box alignments.
\end{enumerate*}
Given a textual phrase and an image as input, the model recognizes the most relevant objects in the image using a pre-trained object detector, and predicts the bounding box referred by the phrase adopting the two aforementioned modules.
Specifically, the rough alignment is based on the similarity score (i.e. concept similarity) between the head of the textual phrase and the predicted label of the bounding boxes. 
Here, the key idea is that the head of the phrase should be very similar (semantically speaking) to the content of the bounding box and, thus, to its class. 


The model is trained to maximize the multimodal similarity between an image and a sentence describing that image, while minimizing the multimodal similarity of the same sentence and a new unrelated image, adequately selected.
We investigated the model performances on the Flickr30k Entities and the ReferIt datasets, showing that our model presents consistent and competitive results in both datasets. 
Moreover, we evaluated our model performance in low-data environments, showing that our model can still achieve surprising results even when trained with just a tiny fraction of training examples

Our contributions can be summarized as follows:
\begin{enumerate*}[label=(\roman*)]
    \item we propose a new model which is based on the novel idea of first predicting a rough alignment between the phrase and a bounding box, and then refining the prediction;
    \item we conduct extensive experiments on the popular Flickr30k Entities and ReferIt datasets, showing state-of-the-art results (in the weakly-supervised setting);
    \item our model, even when trained on a small fraction of the available examples (e.g. 10\%), achieves consistently competitive results.
\end{enumerate*}

\section{Related Works}
\label{sec:sota}
In the literature, there are many works related to our proposal.
Attention was successfully used to generate spacial attention masks able to localize regions referred by phrases preserving linguistic constraints~\cite{xiao2017weakly}, using self-supervision~\cite{javed2018learning}, or even by leveraging a multimodal semantic space~\cite{akbari2019multi}.
Attention was also employed to reconstruct a subset of the input like query subject, location, and context~\cite{liu2019adaptive}, or the full input along with the proposal's information~\cite{chen2018knowledge}. 
Basically, the idea is to reconstruct the input from selected relevant features such that the model learns to ground entities mentioned in the text. 
Along with traditional grounding systems, a regression loss is implemented to refine the bounding boxes coordinates~\cite{liu2021relation}.
Spacial transformer~\cite{jaderberg2015spatial} was successfully employed to compute correlation scores between phrase and image's spacial features map~\cite{zhao2018weakly}. 

Following the encoder-decoder architecture, a slightly different approach consists of learning to ground entity-region by randomly blending arbitrary image pairs, which are reconstructed conditioned by the corresponding texts~\cite{arbelle2021detector}. 
Leveraging the idea of a similarity measure between the two modalities, other works developed a contrastive learning framework where the model localizes entity-region by image-sentence supervision: the contrastive examples may be guided by replacing words in sentences~\cite{gupta2020contrastive}, or either distilling knowledge in order to compute accurate similarity scores~\cite{wang2021improving}.
Using the bounding box's labels and attributes in the representation of image features allows to compute meaningful embedding that can be compared with the language modality~\cite{wang2020maf}. 
Eventually, object detectors may be combined to extract redundant and robust information about regions~\cite{wang2019phrase}. 

A different approach instead overcomes weak supervision by learning to ground through caption-to-image retrieval task: learning caption-to-image retrieval intrinsically means learning to ground, i.e., the model learns to return the correct image given a caption if and only if it has properly understood how to ground the caption with the image~\cite{datta2019align2ground}.

\section{Problem Definition}
\label{sec:problem_definition}
Formally, given an image $\bm{I} \in \mathcal{I}$ and a sentence $\text{S} \in \mathcal{S}$ the VG task aims to learn a map $\gamma: \mathcal{I}\times \mathcal{S}\rightarrow 2^{\mathcal{Q}_\text{S} \times \mathcal{B}_{\bI}}$, where $\mathcal{Q}_\text{S}$ is the domain of the noun phrases defined on S, and $\mathcal{B}_{\bI}$ is the domain of all the bounding boxes defined on $\bI$. More precisely, the set $\mathcal{Q}_\text{S}$ is defined as $\{ \bq_j\}_{j=1}^m$, where $m$ is the number of noun phrases and $\bq_j \in \mathbb{N}^2$ is a vector containing the initial and final character positions in the sentence S.

In this work, given an image $\bI$, we deploy a pre-trained object detector to extract the set of bounding box proposals $\calP_{\bI} = \{( \bc_k, \bh_k, \bl_k)\}^p_{k=1} \subset \mathcal{B}_{\bI}$, where $\bc_k \in \Rset^4$ represents bounding box coordinates, $\bh_k \in \Rset^v$ is the $v$-dimensional vector representing the bounding box features, and $\bl_k \in \Theta$ denotes the class with the highest probability to represent the content of the bounding box over the object detector pre-defined set of categories $\Theta$. 
The bounding box proposal's classification is a common feature offered by most object detectors and will be used in Section~\ref{subsec:similarity-branch} to define the concept similarity.

In the weakly-supervised approach, a training set of $n$ examples is defined as $\mathcal{D} = \{(\bI_i, \text{S}_i)\}_{i=1}^n$. In other words, only the information about sentence $\text{S}_i$ describing the image $\bI_i$ is available at training time, while it is unknown which region $\bb \in \mathcal{B}_{\bI}$ is described by a noun phrase $\bq \in \mathcal{Q}_\text{S}$.
Hence, we learn $\gamma(\bI, S)$ such that it returns a subset \mbox{$\Gamma \subseteq Q_\text{S} \times \calP_{\bI}$} where each couple $(\bq, \bp) \in \Gamma$ aligns the noun phrase $\bq$ to the bounding box proposal $\bp$.

\begin{figure}[t]
  \centering
  \includegraphics[width=0.95\textwidth]{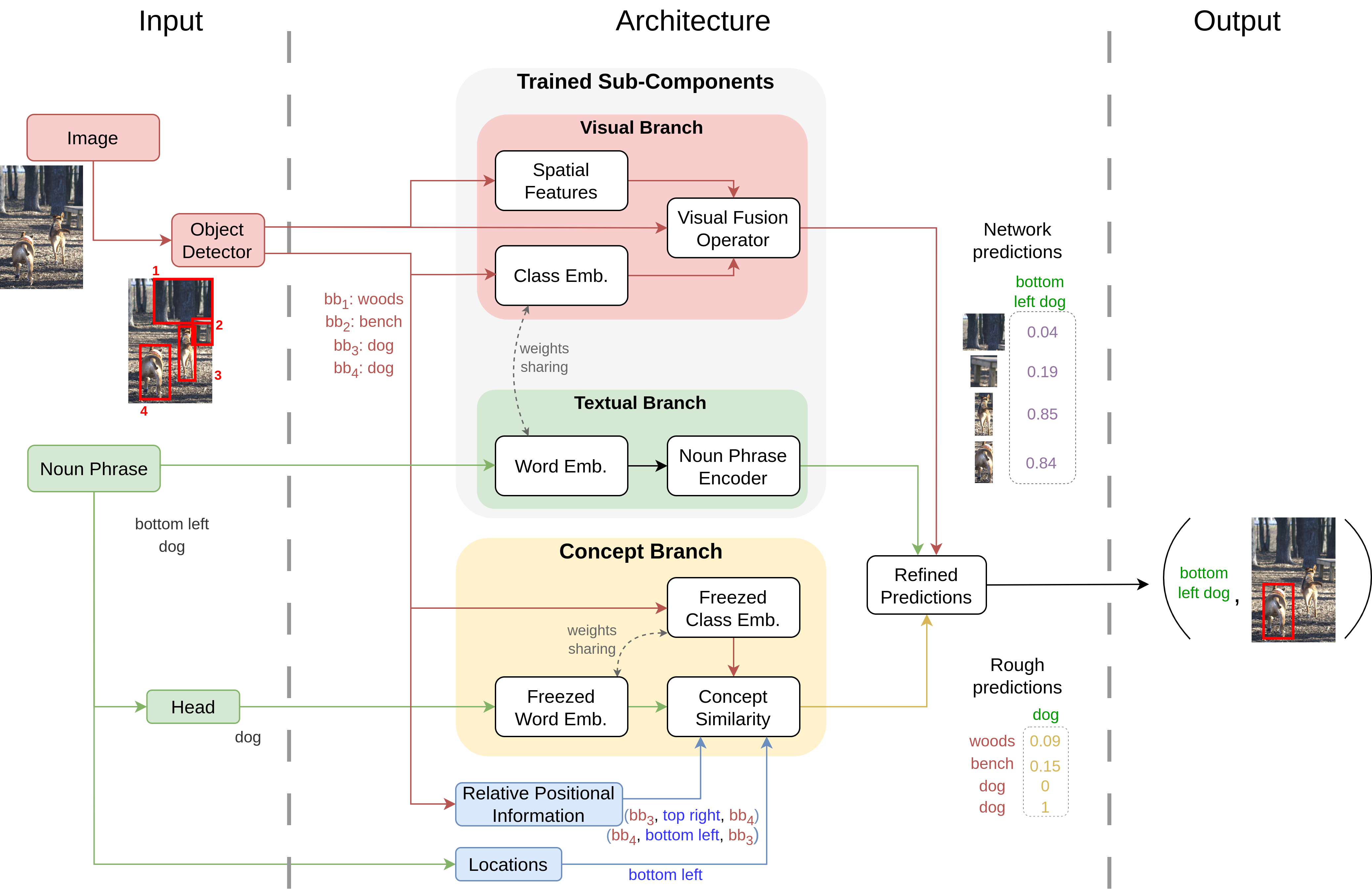}
  \caption{
    \textbf{Our model architecture overview.} The model computes a first rough set of alignments by leveraging prior knowledge from the object detector and word embedding (i.e. \emph{Concept Branch}). 
    A simple positional heuristic is injected as an extra source of prior knowledge to reduce ambiguity for candidate alignments. 
    Then, the visual and textual branches (i.e. \emph{Trained Sub-Components}) match learned multimodal features to predict a second, refined set of alignments. The two sets are then combined together by the \emph{Refined Predictions} module to compute final scores for grounding.
  }
  \label{fig:model-architecture}
\end{figure}

\section{Our Method}
\label{sec:model}
Figure~\ref{fig:model-architecture} depicts our Semantic Prior Refinement Model (SPRM) architecture, which is composed mainly of two modules. 
One is the \emph{Concept Branch} (CB) (see Section~\ref{subsec:similarity-branch}), responsible for predicting a first rough set of region-phrase correspondences. 
Those alignments are obtained through a process named ``concept similarity'' that captures the semantic information conveyed by prior knowledge in object detector and word embedding. 
In particular, it compares the word embeddings of the phrase's head and the bounding box class to get unimodal scores. 
No training is required. 
The information is matched by relying on two important assumptions: 
\begin{enumerate*}[label=(\roman*)]
    \item the proposal's label semantically describes the bounding box content,
    \item and the word embedding space represents the semantic similarity of the words.
\end{enumerate*}
Moreover, the CB includes a positional heuristic that helps to reduce ambiguity for candidate alignments.

The other module (see Section~\ref{subsec:visual-textual_branch}) is made by two sub-components, namely \emph{Visual Branch} and \emph{Textual Branch}, and it is trained to learn a multimodal embedding space for region-phrase correspondences given image-sentence pairs. 
The multimodal representations are constructed to maximize the similarity of region-phrase pairs when both come from the same example, while minimizing the similarity between the regions from the positive example and phrases from another example. 
The second refined set of alignments is obtained by measuring the similarity between learned multimodal visual and textual features for the bounding box proposal and noun phrase. 
The resulting scores are then combined by the prediction refinement module (see Section~\ref{subsec:prediction-module}) to produce final scores. 
The candidate alignment is chosen to be the proposal with maximum similarity with the noun phrase. 


\subsection{Concept Branch}
\label{subsec:similarity-branch}
The \emph{Concept Branch} (CB) is designed to face the most important problem in the weakly-supervised VG: the unavailability of region-phrase ground truths. 
We make use of external sources of knowledge to fill this gap. 
The CB leverages a pre-trained object detector to abstract the content of an image's region through the bounding box classification label, that is the concept expressing the content of the region.
The bounding box classification label is a common feature in most object detectors allowing them to express the content of the bounding box as a concept in the language domain. 
To understand the concept expressed by a textual phrase, we use an off-the-shelf NLP parser deployed to extract the head of the phrase~\cite{javed2018learning}.
In fact, the head of a textual phrase determines its syntactic category. 
Then, by means of a pre-trained word embedding that conveys prior knowledge of words, the CB computes the similarity between the two concepts to obtain a rough score named ``concept similarity''.
In this process, there is \textbf{no training} involved; thus the process is entirely independent of training data and can be treated as prior knowledge.  

This method, although general enough to cover a vast set of cases, suffers from some limitations.
First, the proposal's classification may be noisy and incorrect, driving the CB to inaccurate alignments. 
Second, the word embedding similarity may be biased and imprecisely captures the semantic similarity between words. 
Third, the CB produces equal scores when proposals have the same label. 
In order to deal with this issue, we adopted another source of prior knowledge based on spatial relations. 
For proposals with the same label, we extract relative positional information (e.g. ``top'').
We then match those relations with a location extracted from the phrase by a simple text search (e.g. ``left'' in ``dog on the left'').

Formally, given a set of $p$ bounding box proposals $\calP_{\bI}$, let $\EPI = \{ \ePI_k \}^{p}_{k=1}$ be the corresponding set of $g$-dimensional vectorial embeddings, where each $\ePI_k$ is the embedding of the bounding box class $\bl_k$, for $1\leq k \leq p$. 
Given a noun phrase $\bq_j$ composed by a sequence of $L(\bq_j)$ words $W^{\bq_j} = [ w^{\bq_j}_1 \ldots w^{\bq_j}_{L(\bq_j)}]$, let $\Eqj = \{ \eqj_i \}^{L(\bq_j)}_{i=1}$ 
be the set of words embedding of size $g$ associated with each word in the noun phrase $\bq_j$. 
Let $\bs^t_j \in \Rset^6$ and $\bs^v_k \in \Rset^6$ be two multi-hot vectors that encode locations in $\bq_j$ and relations in the $k$-th proposal, respectively. 
Then, the concept similarity score for each proposal is:
\begin{align*}
  \bS_{jk}  &= f_{mask}\left(\erqj_j, \ePI_k, \bs^t_j, \bs^v_k \right) 
            =\begin{cases} 
                   \fsim \left( \erqj_j, \ePI_k \right)         & \text{if } \left(\bs^v_k\right)^\top \bs^t_j \ge 0 ; \\
                  -1                                            & \text{otherwise.}
            \end{cases}\nonumber
\end{align*}
where $\fsim$ is a similarity measure (e.g. cosine similarity) and $\fsim ( \erqj_j, \ePI_k )$ returns the similarity score between word embeddings of phrase's head $\erqj_j$ and proposal's label $\ePI_k$. The function $f_{mask}$ returns this similarity score only when the phrase and the proposal share at least one spatial reference, otherwise $-1$ is returned.

\subsection{Visual and Textual Branches}
\label{subsec:visual-textual_branch}
Given the set of bounding box proposals $\calP_{\bI}$ detected in the image $\bI$ by the object detector, for each of them, our model extracts the spatial features $H^s = \{ \bh^s_k \}^p_{k=1}$ where $\bh^s_k \in \Rset^5$, as indicated in~\cite{DBLP:conf/sac/RigoniSS22}. 
Moreover, contrary to the \emph{Concept Branch}, the Visual and Textual branches adopt trainable word embeddings $\EPIt = \{ \ePIt_k \}^{p}_{k=1}$ and $\Eqjt = \{ \eqjt_i \}^{L(\bq_j)}_{i=1} $ associated to the bounding box classes and to the words of the noun phrases, respectively.

Initially, both visual and spatial features are concatenated and then projected on a smaller dimensional space, thus leading to a set of new vectorial representations $H^{||} = \{\bh^{||}_{k} \}^p_{k=1}$, with  \(\bh^{||}_{k} = \bm{W}^{||} \big( \bh^s_k || \bh_k \big) + \bm{b}^{||}\), 
where $||$ indicates the concatenation operator, \mbox{$\bh^{||}_{k} \in \Rset^g$}, $\bm{W}^{||} \in \Rset^{g \times (5 + v)}$ is a matrix of weights, and $\bm{b}^{||} \in \Rset^g$ is a bias vector.
The new representation is then summed to the word embedding of the bounding box label to obtain the final visual features \( \bh^v_{k} = \bh^{||}_{k} + \ePIt_k \), 
where $\bh^v_{k} \in \Rset^g$.

Given the set $\Eqjt$ of trainable word embeddings associated with the noun phrase $\bq_j$, the textual branch applies a function $f_{enc}$ to generate only one embedding $\bh^t_j\in \Rset^\tau$ for each phrase $\bq_j$. 
This textual features extraction is defined as \( \bh^t_j = f_{enc}(\Eqjt) \).

Note that the embeddings $\EPIt$ and $\Eqjt$ are generated with trainable modules that share the weights among each other (weights sharing). 
So, during training, the word embeddings learn multimodal embeddings for the visual and textual information.

\subsection{Refined Predictions}
\label{subsec:prediction-module}
The prediction module is in charge of refining the rough predictions $\bS_{jk}$, i.e., the \emph{Concept Branch} predicted scores, using the visual $\bh^v_k$ and textual $\bh^t_j$ features.
Initially, starting from $\bh^v_{k}$ and $\bh^t_{j}$, the model predicts the probability $\bP_{jk}$ that a bounding box proposal of index $k$  is referred by the noun phrase $\qj$ as \( \bP_{jk} = \fsim ( \bh^v_{k} , \bh^t_j ) \),
where $\fsim$ is a similarity measure between vectors. 
Please note that in our work, we adopt the cosine similarity function; therefore, $\bh^v_{k}$ and $\bh^t_{j} $ have the same vector dimension, i.e. $g = \tau$.
 
Finally, the rough predictions are refined in $\bm{\hat{P}}_{jk} = \omega * \bP_{jk} + (1-\omega) * \bS_{jk}$ using an hyper-parameter \mbox{$\omega \in \{x \in \Rset \mid 0 \leq x \leq 1\}$}.
Therefore, the model predictions are not constrained to values defined by concept similarity, but they co-work for the final predictions.

\subsection{Loss Function}
Inspired by~\cite{wang2020maf}, we adopt a contrastive loss. 
The contrastive objective $\calL$ aims to learn the visual and textual features by maximizing the similarity score between paired image-sentence examples and minimizing the score between the negative examples.

Formally, given two training examples $(\bI, \text{S}), (\bI', \text{S}') \in \mathcal{D}$ such that $\text{S} \neq \text{S}'$ and $\bI \neq \bI'$, the loss function $\calL$ is defined as:
\begin{align*}
  &\calL =  - \underbrace{f_{pair}(\bI, \sent)}_\text{Positive example} + \underbrace{f_{pair}(\bI', \sent)}_\text{Negative example}, 
  &f_{pair}(\bI, \sent) = \frac{1}{m} \sum^m_{j=1} \max_k \frac{\bm{\hat{P}}_{jk}}{\sum^p_i \bm{\hat{P}}_{ji}},
\end{align*}
where $f_{pair}$ is the similarity function defined over the multimodal pair image-sentence, $m$ is the number of queries in $\sent$ and $\bm{\hat{P}}_{jk}$ is the predicted similarity between noun $\bq_j$ and proposal $\bm{p}_k$.
Basically, the goal of $f_{pair}$ is to aggregate the similarity scores of all the region-phrase pairs, determining the degree to which the phrases correspond with the content of the image.

In contrast to what is done in~\cite{wang2020maf} where for each positive example, several negative examples built from the batch are considered, we adopt just a specific negative example $(\bI', \text{S})$.
The negative example is built from the example $(\bI', \text{S}')$, selected from the batch precisely to be the one where the sentence $\text{S}'$ is the most similar to the sentence $\text{S}$. 
This allows the model to focus on fine-grained region-phrase details that differ between the two examples.
Precisely, given a training example $(\bI, \text{S}) \in \mathcal{B}$, the negative example is chosen as:
\begin{align*}
    &(\bI', \text{S}') = \argmax_{(\bI'', \text{S}'') \in \mathcal{B}'}
    f_{sim}(\zeta(\text{S}''), \zeta(\text{S})) ,
    &\zeta(\text{S}) = \frac{1}{m} \sum^m_{j=1} \frac{1}{L(\bq_j)} \sum^{L(\bq_j)}_{i=1} \eqj_i
\end{align*}
where $\mathcal{B}' = \mathcal{B}\backslash\{(\bI, \text{S})\}$.
Thus, the similarity is measured in the word embedding space.

\section{Experiments}
\label{sec:esperimental_assessment}

\subsection{Datasets and Evaluation Metrics}
In this work, we have evaluated our model on the Flickr30k Entities~\cite{plummer2015flickr30k} and ReferIt~\cite{kazemzadeh2014ReferItgame} datasets.\footnote{We considered the two most largely adopted datasets among the $15$ papers used as comparison in our work. }
The Flickr30k Entities dataset contains $32$K images and $360$K queries, while the ReferIt~\cite{kazemzadeh2014ReferItgame} dataset contains $20$K images and $120$K queries.
Following the previous works in the area, we adopted \emph{Accuracy} as the main evaluation metric.
Namely, given a noun phrase, it considers a bounding box prediction to be correct if and only if the {\it Intersection over Union} value between the predicted bounding box and the ground truth bounding box is at least 0.5.
Moreover, we also calculate the \emph{Pointing Game Accuracy} for comparison purposes~\cite{akbari2019multi,datta2019align2ground,ramanishka2017top,liu2021relation}.
\emph{Pointing Game Accuracy} considers an example to be positive whether the center of the predicted bounding box lies wherever inside the ground-truth box. 

\subsection{Model Selection and Implementation} 
\label{sec:model_selection_implementation_details}
The model selected for evaluating the test set of the Flickr30k Entities and of the ReferIt datasets is chosen on the epoch that better performs in terms of \emph{Accuracy} in the validation set. 
We search for the best hyper-parameters on both Flickr30k Entities and ReferIt datasets, independently on the considered fractions of training data $\{5\%, 10\%, 50\%, 100\%\}$ used for learning.
We selected $10^{-5}$ as the learning rate step and we have adopted GloVe~\cite{pennington2014glove} as word embeddings, where $\tau = g = 300$.
In our work, $f_{enc}$ is implemented with a LSTM~\cite{schmidhuber1997long} neural network.
The vector $\bh^t_j$ is the $\tau$-dimensional LSTM output of the last word $w^{\bq_j}_{L(\bq_j)}$ in the noun phrase $\bq_j$.
The bounding box proposals $\calP_{\bI}$ are extracted with the Bottom-Up Attention~\cite{anderson2018bottom} object detector with a confidence score of $0.1$ for Flickr30k\footnote{We used the same features of~\cite{wang2020maf}.} Entities and $0.2$ for ReferIt\footnote{\url{https://github.com/MILVLG/bottom-up-attention.pytorch}}.
The bounding box features have a dimension of $v = 2048$.
We use the cosine similarity as a similarity measure $\fsim$ between vectors.
Our SPR model code is publicly available online\footnote{\url{https://github.com/drigoni/SPRM/}}.

\subsection{Experimental Results}
We compare our model to several approaches in the literature on the Flickr30k Entities and ReferIt datasets. 
We also assess our model performance when trained only with a small number of training examples.
Indeed, the untrained \emph{Concept Branch} module should give stability to the model even when it is trained on a small training set, as it should help to counter the overfitting trend that occurs with small datasets.

\subsubsection{Full Training Set Scheme}
\label{subsec:full-training-scheme}
\begin{table}[t]
  \centering
  \footnotesize
  \begin{tabular}{@{\hspace{0\tabcolsep}}l@{\hspace{1\tabcolsep}}c@{\hspace{1\tabcolsep}}c@{\hspace{1\tabcolsep}}c@{\hspace{1\tabcolsep}}cc@{\hspace{1\tabcolsep}}c@{\hspace{0\tabcolsep}}}
    \toprule
    \multirow{2}{*}[-2pt]{Model} & Backbone & Proposals   &\multicolumn{2}{c}{Flickr30k E. (\%)} & \multicolumn{2}{c}{ReferIt (\%)} \\
    \cmidrule(rl{4pt}){4-5} \cmidrule(rl{4pt}){6-7}
    & (Pre-training) & (Pre-training) & $\uparrow$ Acc. &$\uparrow$ P. Acc. &$\uparrow$ Acc. &$\uparrow$ P. Acc. \\
    \midrule
    Top-down Saliency~\cite{ramanishka2017top}                      &InceptionV3 (IN)       &-              &-          &$50.10$     &-          &-       \\
    KAC Net~\cite{chen2018knowledge}                                &VGG16 (VOC)            &SS,EB          &$38.71$     &-          &$15.83$     &-       \\
    Semantic Self-Sup.~\cite{javed2018learning}                     &VGG16 (IN)             &-           &-          &$49.10$     &-          &$39.98$  \\
    Anchored Transformer~\cite{zhao2018weakly}                      &VGG16 (VOC)            &EB       &$33.10$     &-          &$13.61$     &-       \\
    Multi-level Multimodal~\cite{akbari2019multi}                   &PNASNet                &-           &-          &$69.19$    &-          &$48.42$ \\
    Align2Ground~\cite{datta2019align2ground}                       &RN152 (IN)             &BUA (VG)     &-          &$71.00$     &-          &-       \\
    Counterf. Resilience~\cite{fang2019modularized}                 &RN101 (IN)             &F-RCNN (CC)      &$48.66$    &-          &-          &-       \\ 
    MAF~\cite{wang2020maf}                                          &RN101 (IN)             &BUA (VG)      &\underline{$61.4$}     &-          &-          &-       \\
    Contrastive Learning~\cite{gupta2020contrastive}                &RN101 (IN)             &BUA (VG)      &$51.67$    &$76.74$     &-          &-       \\
    Grounding By Sep.~\cite{arbelle2021detector}                    &VGG16, PNASNet (IN)    &-      &-          &$75.60$    &-          &$58.21$    \\      
    Relation-aware~\cite{liu2021relation}                           &RN101                  &F-RCNN (VG)      &$59.27$    &$78.60$    & $37.68$   &$58.96$    \\
    Contrastive KL Distill.~\cite{wang2021improving}                &RN101 (IN)             &BUA (VG)      &$53.10$    &-          &$38.39$    &-          \\
    EARN~\cite{liu2022entity}                                       &RN101                  &EB, F-RCNN      &$38.73$    &-          &$36.86$    &-          \\
    RefCLIP~\cite{jin2023refclip}                                   &Darknet-53             &YoLo3 (VG)      &-          &-          & $42.64$   &-          \\
    SimMaps~\cite{shaharabany2023similarity}                        &VGG16 (IN)             &-      &$45.56$    &$79.95$    & $38.74$   & $70.25$   \\
    \midrule
    SPR baseline + CLIP (ours)                                      &RN101 (IN)             &BUA (VG)       &$56.89$    &$77.06$    & $40.99$   & $57.48$   \\
    \textbf{SPR model (ours)}                                       &RN101 (IN)             &BUA (VG)       &$\bm{62.20}$  &$\bm{80.68}$ &$\bm{48.04}$ &$\bm{62.40}$ \\
    \bottomrule
  \end{tabular}
  \caption{\textbf{Results on Flickr30k Entities and ReferIt test sets}. \emph{Acc.} is the standard accuracy metric, while \emph{P. Acc.} is the pointing game accuracy metric. For each work we report in the column \emph{Backbone} the adopted visual features encoder, abbreviating ResNet-101/152~\cite{DBLP:conf/cvpr/HeZRS16} with RN101/152. In the \emph{Proposals} column we listed proposal networks or object detectors employed, where SS: Selective Search~\cite{uijlings2013selective}, EB: EdgeBox~\cite{zitnick2014edge}, F-RCNN: Fast-RCNN~\cite{ren2015faster}, BUA: Bottom-up Attention~\cite{anderson2018bottom}. In both columns, the pre-training dataset is indicated in parenthesis, whenever available, following these abbreviations: VG: Visual Genome~\cite{krishnavisualgenome}, IN: ImageNet~\cite{imagenet_cvpr09}, CC: MS-COCO~\cite{lin2014microsoft}, VOC: PASCAL VOC~\cite{everingham2010pascal}.
  }
  
  
  \label{tab:table_all_results}
\end{table}
Table~\ref{tab:table_all_results} compares our model results to those of several approaches in the literature.
Our model proposal outperforms all other approaches on standard \emph{Accuracy} and \emph{Pointing Game Accuracy}. 
In particular, in the Flickr30k Entities, our model's improvements over the State-of-the-Art are $+0.8\%$ in Accuracy and  $+2.08\%$ in P. Accuracy.
While on ReferIt, the improvements are $+9.65\%$ and $+3\%$, respectively for both the metrics.

To assess the soundness of our approach we tested a variant of our model that replaces visual and textual branches, responsible to learn the multimodal embedding space, with CLIP's multimodal embeddings (referred as SPR baseline + CLIP)~\cite{radford2021learning}. 
As the results show, in Table~\ref{tab:table_all_results}, our full SPR model still outperforms the variant with CLIP. 
This occurs because CLIP was trained to capture the multimodal coarse-grained information from image and sentence pairs, while in VG we need more fine-grained details regarding the alignments region-query.

\begin{figure}[t]
\centering
\begin{minipage}[t]{.49\linewidth}
    \centering
    \begin{tikzpicture}[scale=0.73]
    \footnotesize
    \begin{axis}[
        width=9cm,
        height=6cm,
        legend style={at={(0.5,1.01)},anchor=south,draw=none},
        legend columns=-1,
        xlabel={$\omega$\rule{0pt}{7pt}},
        xmin=0.1, xmax=0.9,
        ymin=43, ymax=65,
        xtick={0.1,0.25,0.4,0.5,0.75,0.9},
        grid=major,
        grid style={dashed},
        hide y axis,
        point meta=explicit symbolic,
        nodes near coords
        ]
    \addplot[smooth,mark=*,color=cyan,text=black] 
    plot coordinates {
        (0.10,59.63) [59.63\%]
        (0.25,61.66) [61.66\%]
        (0.40,62.20) [62.20\%]
        (0.50,61.51) [61.51\%]
        (0.75,57.06) [57.06\%]
        (0.90,55.99) [55.99\%]
    };
    \addlegendentry{Flickr30k Entities}

    \addplot[smooth,color=orange,mark=x,text=black]
    plot coordinates {
        (0.10,44.63) [44.63\%]
        (0.25,46.57) [46.57\%]
        (0.40,47.58) [47.58\%]
        (0.50,47.96) [47.96\%]
        (0.75,48.04) [48.04\%]
        (0.90,45.61) [45.61\%]
    };
    \addlegendentry{ReferIt}
    \end{axis}
\end{tikzpicture}\vspace{-8pt}
    \caption{\textbf{Accuracy results on Flickr30k Entities and ReferIt test sets varying the $\omega$ hyper-parameter}. Results obtained by training the model on $100\%$ of the training set.}
    \label{fig:plot_omega}
\end{minipage}
\hfill
\begin{minipage}[t]{.49\linewidth}
    \centering
    \begin{tikzpicture}[scale=0.73]
    \footnotesize
    \begin{axis}[
        width=9cm,
        height=6cm,
        legend style={at={(0.5,1.01)},anchor=south,draw=none},
        legend columns=-1,
        xlabel=Training Fraction (\%),
        xmin=0, xmax=100,
        ymin=40, ymax=65,
        xtick={0,5,10,50,100},
        grid=major,
        grid style={dashed},
        hide y axis,
        point meta=explicit symbolic,
        nodes near coords
        ]
    \addplot[smooth,mark=*,color=cyan,text=black] 
    plot coordinates {
        (005,57.20) [57.20\%]
        (010,60.72) [60.72\%]
        (050,61.49) [61.49\%]
        (100,62.20) [62.20\%]
    };
    \addlegendentry{Flickr30k Entities}

    \addplot[smooth,color=orange,mark=x,text=black]
    plot coordinates {
        (005,41.02) [41.02\%]
        (010,42.67) [42.67\%]
        (050,47.68) [47.68\%]
        (100,48.04) [48.04\%]
    };
    \addlegendentry{ReferIt}
    \end{axis}
\end{tikzpicture}\vspace{-8pt}
    \caption{\textbf{Accuracy results on Flickr30k Entities and ReferIt test set by our model trained in low-data environments}. The percentage refers to the fraction of the training set considered during training.}
    \label{fig:plot_training_size}
\end{minipage}
\end{figure}

The hyper-parameter $\omega$ regulates the weight of the \emph{Concept Branch} on the final predictions: the higher the value, the less the \emph{Concept Branch} affects final predictions.
For this reason, in Figure~\ref{fig:plot_omega} we present the \emph{Accuracy} results obtained with our model trained on the entire training set at different values of $\omega$: $\{0.1, 0.25, 0.4, 0.5, 0.75, 0.9\}$.
As shown by the chart, $\omega$ greatly affects the model performance in both datasets, allowing the model to reach its peak of performance when $\omega=0.4$ in Flickr30k Entities and $\omega=0.75$ in ReferIt.

\subsubsection{Small Training Set Scheme}
In this section, we present the results obtained with our model on the datasets where only a fraction of training examples are used for training. 
Figure~\ref{fig:plot_training_size} reports our model \emph{Accuracy} results.
On Flickr30k Entities, the model is able to obtain State-of-the-Art results even when trained with only 50\% of the training data, while on ReferIt, even when the model is trained with 5\% of the training examples, it achieves State-of-the-Art performances.
As expected, the \emph{Concept Branch} module, which does not require training, makes the model training more stable and helps to counter the overfitting trend that occurs with small datasets.

\subsubsection{Model Ablation}
\label{sec:ablation}
In this section, we assess the performance of our model's components:
\begin{enumerate*}[label=(\roman*)]
    \item the untrained \emph{Concept Branch},
    \item the trained visual and textual branches,
    \item and the \emph{Relative Positional Information} component.
\end{enumerate*}
The model achieves the best results when both the \emph{Concept Branch} and the trained modules jointly work to produce the final predictions, as shown in Table~\ref{tab:table_ablation}.

\begin{wraptable}{r}{0.52\linewidth} 
  \centering
  \footnotesize
  \begin{tabular}{c@{\hspace{1\tabcolsep}}c@{\hspace{1\tabcolsep}}c@{\hspace{1\tabcolsep}}c@{\hspace{1\tabcolsep}}c@{\hspace{1\tabcolsep}}}
    \toprule
    Concept & Trained & Rel. Posit. & Flickr30k  &ReferIt \\ 
    Branch  & Modules & Information & Entities (\%) &(\%) \\
    \midrule
    \XSolidBold     &\CheckmarkBold &\XSolidBold     &$23.52$        & $15.03$   \\
    \CheckmarkBold  &\XSolidBold    &\XSolidBold     &$54.96$        & $40.07$   \\
    \CheckmarkBold  &\XSolidBold    &\CheckmarkBold  &$55.02$        & $42.69$   \\
    \CheckmarkBold  &\CheckmarkBold &\XSolidBold     &$62.10$        &$45.44$    \\
    \CheckmarkBold  &\CheckmarkBold &\CheckmarkBold  &$\bm{62.20}$   &$\bm{48.04}$ \\
    \bottomrule
  \end{tabular}
  \caption{\textbf{Model Ablation}. Accuracy of our model's components. The \emph{Concept Branch} contributes more to the final model performances.}
  \vspace{-5pt}
  \label{tab:table_ablation}
\end{wraptable}
The boost in \emph{Accuracy} given by the \emph{Concept Branch} is significant: $+38.58\%$ and $+30.41\%$ for Flickr30k Entities and ReferIt, respectively.
As expected, the \emph{Relative Positional Information} component constantly improves the model accuracy by $+0.1\%$ on Flickr30k Entities and by $+2.6\%$ on ReferIt.
Further investigations showed that Flickr30k presents few spatial references in the queries, which explains the difference in performance gains between the two datasets. 

\subsubsection{Comparison to V\&L models}
The recent success of CLIP~\cite{radford2021learning} in learning image-level visual representations from image-text pairs has inspired a new line of research~\cite{Li_2022_CVPR,zhong2022regionclip} to extend large V\&L models on fine-grained correspondence between sentences and objects in images. 
In the same direction of research, in this section we compared our model to GLIP~\cite{Li_2022_CVPR}. 
GLIP aims to learn region-level visual representations, thus enabling fine-grained alignment between image regions and textual concepts and works in a \textbf{fully-supervised fashion}. It is trained on $27$M of grounding data, including Flickr30k. All the ground alignments are used, when available, during training. Thus the comparison between our model, which uses weak annotations, and GLIP is unfair. Nevertheless, we compared the two methods in the zero-shot setting, i.e. GLIP-T (B) trained only on object detection dataset Objects365 against our \textbf{untrained} Concept Branch (CB). GLIP-T (B) obtains $36.10\%$ accuracy on Flickr30k while our CB scores $55.02\%$.

\subsection{Limitations}

Our model limitations stem mainly from the word embedding and the object detector components.
In fact, the design of our proposal is well-suited for GloVe, Bottom-Up Attention, and LSTM components.
However, these approaches are no more State-of-the-Art.
Modern approaches, such as Large Language Models (LLMs) like BERT~\cite{devlin2018bert}, could improve the performance of our model. 
Indeed, LLMs take advantage of their effective contextual capabilities to embed words in a sentence.
In our architecture, LLMs can replace:
\begin{enumerate*}[label=(\roman*)]
    \item the LSTM in the \emph{Textual Branch}, and
    \item the current GloVe embeddings in the \emph{Concept Branch}.
\end{enumerate*}
In both cases, the introduction of this new component is not straightforward, especially in the \emph{Concept Branch}.
In fact, the concept similarity scores are computed between the head of the phrase and bounding box classes. 
Thus, it is not clear what context the LLMs should consider during the embedding of class labels.

Furthermore, our model's dependency on the object detector performance is made explicit by the Concept Branch. Usually, in other works this dependency is hidden in the multimodal features fusion process, which relies on visual features and proposals from the object detector's output. 
To ensure an objective comparison of results, we use the same object detector that the current State-of-the-Art model MAF~\cite{wang2020maf} utilizes. This detector is Faster-RCNN with ResNet-101, which has been trained on Visual Genome. We also make sure to adopt precisely the same features shared by MAF's authors.
However, more recent object detectors could improve our model's performance.

\section{Conclusion}
\label{sec:conclusion}
Our work focused on tackling the task of weakly-supervised visual-textual grounding, where the lack of ground truth alignments presents a challenge for learning. 
Our core contribution resides in the \emph{Concept Branch}. It captures the semantic similarity between the image's region and the phrase by matching the bounding box class and the phrase's head in a word embedding space. 
The alignments are obtained leveraging pre-trained object detector and word embedding, thus training is not required. 
The new knowledge does not depend on training data and can be treated as prior with many advantages. 
First, it enables compositionality as new models could be built on top of the prior to avoid starting from scratch. 
Second, this knowledge helps the training phase, especially in the first epochs where the model can't be guided as in the fully-supervised setting. 
Third, the independence of training data also enables performance stability which makes our model suitable in low-data environments.
As proven by our results, this approach presents State-of-the-Art performance on Flickr30k Entities and ReferIt benchmarks. 
Inspired by~\cite{liu2021relation}, future works aim to extend our loss function to include a bounding box regression component, that has been proven to boost VG models performances. Additionally, future work will explore the use of different object detectors' categories~\cite{rigoni2023cleaner}. Finally, inspired by~\cite{dost2020vtkel}, we aim to incorporate knowledge graph information in the model, enhancing the \emph{Concept Branch} module with more structured information.

\paragraph{Acknowledgements.}
We acknowledge the support of the PNRR project FAIR - Future AI Research (PE00000013), under the NRRP MUR program funded by the NextGenerationEU. This research was also supported by an UniPD BIRD-2021 Project and by the PRIN-17 PREVUE project from the Italian MUR (CUP E94I19000650001).
We finally acknowledge EuroHPC Joint Undertaking for awarding us access to Vega at IZUM, Slovenia. 

\bibliography{biblio}
\end{document}